\documentclass{article}
\usepackage{ijcai19}

\usepackage{times}
\usepackage{soul}
\usepackage{url}
\usepackage[hidelinks]{hyperref}
\usepackage[utf8]{inputenc}
\usepackage[small]{caption}
\usepackage{graphicx}
\usepackage{amsmath}
\usepackage{algorithm}
\usepackage{algorithmic}
\usepackage{comment} 
\usepackage{apacite}
\usepackage[]{todonotes}
\usepackage{booktabs}
\usepackage[markup=underlined]{changes}

\title{Let's Make It Personal \\ A Challenge in Personalizing Medical Inter-Human Communication}

\author{
Mor Vered$^1$
\and
Frank Dignum$^2$\And
Tim Miller$^1$
\affiliations
$^1$School of Computing and Information Systems, The University of Melbourne\\
$^2$Department of Information and Computing Sciences, Universiteit Utrecht\\
\emails
\{mor.vered, tmiller\}@unimelb.edu.au,
f.p.m.dignum@uu.nl
}

\begin{document}

\maketitle

\vspace{-40pt}

\begin{abstract}

\vspace{-8pt}

Current AI approaches have frequently been used to help personalize many aspects of 
medical experiences and tailor them to a specific individuals' needs. 
However, while such systems consider medically-relevant information, they ignore socially-relevant information about \emph{how} this diagnosis  should be communicated and discussed with the patient. The lack of this capability may lead to mis-communication, resulting in serious implications, such as patients opting out of the best treatment. Consider a case in which the same treatment is  proposed to two different individuals. The manner in which this treatment is mediated to each should be different, depending on the individual patient's history, knowledge, and mental state. While it  is clear  that this communication should be conveyed via a human medical expert and not a software-based system, humans are not always capable of considering all of the relevant aspects and traversing all available information. We pose the challenge of creating Intelligent Agents (IAs) to assist medical service providers (MSPs) and consumers in  establishing a more personalized human-to-human dialogue. Personalizing conversations will enable patients and MSPs to reach a solution that is best for their particular situation, such that a relation of trust can be built and commitment to the outcome of the interaction is assured. 
We propose a four-part conceptual framework for personalized social interactions, expand on which techniques are available within current AI research and discuss what has yet to be achieved. 
    
\end{abstract}

\maketitle

\vspace{-8pt}

\section{Introduction}
Generating personalized, tailored assistance is an important topic in AI research. From Intelligent Tutoring Systems, which attempt to discern between different learning techniques \shortcite{gal2015making}, to tailored negotiation schemes \shortcite{kraus1998reaching
}, AI plays an increasingly prominent role by either providing important training, providing people with best practice advice or acting on people's behalf. The transition to electronic medical records and  availability of patient data has also lead to an increase of AI within the medical domain assisting qualified health providers with both personalized diagnoses and  treatment suggestions \shortcite{dilsizian2014artificial}.  As of yet most of this support, while personalized in content, is conveyed to the user as a standard template advice. 
There is no support on the \emph{manner} in which output should be communicated and discussed between people, based on socially-relevant criteria. Existing work from behavioral economics further emphasizes the implications of the existing methods by   demonstrating that the manner in which treatment suggestions have been conveyed has a major impact on the likelihood that the patients will comply with the prescribed course of treatment \shortcite{emanuel2016using}. 

Upon proposing and describing a required treatment, an MSP cannot be expected to refer, on a personal level, to each of her  numerous patients individual criteria, among which are culture, age, socioeconomic status, mental state, marital status, personal history, and even personality. Consider the example of a young, 22-year-old, new mother with a high school diploma as opposed to a single 56-year-old male, professor of computer science, both of whom have been diagnosed with the same condition. The manner in which diagnosis and treatment are described should be different to each individual, and perhaps different again if the patient is joined by their family. These effect not only the content of the explanation, 
but also 
the tone, manner and language used.

While we cannot expect a MSP to know and consider all of these variables, the implications of generic advice delivery may be considerable and even destructive, resulting in refusal of treatment and non-compliance to medication, even for life saving treatments \shortcite{ito2017qualitative,
puts2010characteristics}. 
\emph{Trust} between provider and consumer has been shown to be one of the core reasons for complying with medical treatment \shortcite{penman1984informed}. We seek a higher level of personalized communication as a means for increasing trust and therefore treatment acceptance and compliance.   
 
Intelligent Agents (IAs) can act as a supporting entity for doctors, aiding people in creating a more personalized dialogue. While doctors use their expertise and knowledge to determine the best course of action, an IA could help translate this information to a patient in the best possible manner while considering relevant patient information and alleviating the work load of the doctor in the process. 
 For example, additional information could indicate that a particular course of treatment might be intimidating to the patient due to being un-affordable for current socioeconomic status. In this instance the medical professional might want to emphasize the repercussions of not conforming to the suggested treatment and alert the patient with regards to financial aid possibilities. 

The current flourish of AI research and applications can enable us to make this grand challenge a reality. Our community has encouraged and explored interdisciplinary research connecting computer science with the behavioral and social sciences, learning about human mental models, and explainable AI. 
We have explored the complementary fields of negotiation \shortcite{aydougan2018machine}, personalized interaction  \shortcite{nooitgedagt2017coaching}, persuasion \shortcite{moulin2002explanation}, explanation from the human perspective \shortcite{miller2018insights,madumal2018towards} as well as a recent increase in the applicability of text mining \shortcite{kocbek2016text}, machine learning and deep learning algorithms \shortcite{erickson2017machine
}. By building on this foundational research, we can assist doctors and patients  to create a more personalized inter-human communication.
We will present a conceptual framework to realize the proposed system, discuss what is already done for each component and what still remains to explore as exciting, new, research directions for the AI community.

\vspace{-8pt}

\section{Conceptual Framework Design}

\begin{figure}[h]
  \includegraphics[width=\linewidth]{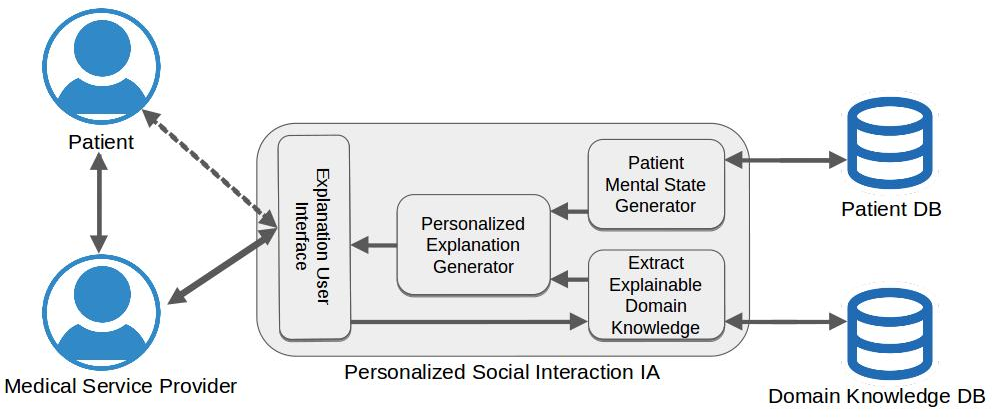}
  \caption{A conceptual design for a Personalized Social Interaction Framework.}
  \label{fig:framework}
\end{figure}

We begin by breaking the task down into four  main components, each component presenting a hard problem in the field of artificial intelligence.  In this manner each of these components can, initially, be addressed separately. Figure~\ref{fig:framework} presents our conceptual design 
for a \emph{Personalized Social Interaction Framework} with the arrows presenting the flow of information between the different components.

\vspace{-8pt}

\paragraph{Extract Explainable Domain Knowledge}

The first component addresses the IA's ability to extract the relevant domain knowledge which may be used to generate a general explanation. To achieve this goal the component must have access both to the MSP's expert recommendation (perhaps also given by an algorithm) and to a database of the required domain knowledge. This will enable the IA  to sift through the data  and determine which domain knowledge information will be relevant and may help construct a general explanation. At this point the explanation is not yet personalized to a specific consumer. For example, in the medical domain, this would mean that two different individuals diagnosed with the same medical condition would be given the same general explanation. Notice that the flow of information between the component and the database is bi-directional. This will allow the updating of the database when coming across  new, relevant information. For example, when  encountering a new symptom associated with a known disease. 

Much work has been done towards achieving a more tailored approach  to  medical  related  procedures.   Powerful  and  accurate text mining techniques and advanced analysis of "big data" in health care can potentially provide diagnoses based on empirical data. The transition to electronic health records (EHRs)  has  made  the  availability  of  patient  data  a  reality. Partnered with increasing access to high-performance computing systems we are becoming increasingly closer to being able to support and sustain personalized medical treatment. Text mining techniques, taking advantage of information derived from multiple data sources such as radiology reports, pathology reports and patient and hospital admission data, and are able to classify these existing clinical records to specific diseases \shortcite{kocbek2016text};

 There has also been research in personalizing the interpretation of information from multiple, disjoint sources.  \shortciteA{amir2016intelligent} presents the problem of information sharing in loosely-coupled teams, such as treating patients with several different carers, co-authoring documents and developing soft-ware products. In these cases, additional information must be extracted  from  a  large,  shared,  information  database  in  order to form an appropriate decision.  
 
 Clinical Decision Support (CDS) systems aim to help promote informed decision making and improve patient safety. These systems are designed to assist physicians and other health professionals with clinical decision-making tasks aiming to provide assessments, classification and concrete recommendations for care \shortcite{greenes2018clinical}. 

Combined together, the above methods have the ability to aid in successfully personalizing treatments and diagnoses of individual patients however they have not focused on the \emph{manner} in which said treatment should be conveyed to the patient. None of these methods have investigated  interaction with patients and the information needed to facilitate such an interaction. 
Considering the social interaction would directly  effect which information would be deemed relevant for the patient. Relevant no longer only pertains to medically relevant, but also \emph{socially relevant}.

\paragraph{Patient Mental State Generator} 
This component addresses the specific needs of the patient by attempting to build a 
mental model of the patient using knowledge obtained from Theory of Mind. 
    The generated mental model does not need to address the specific context of the conversation or the MDP's expert recommendation but rather is built as a general mental model of the current state of the patient considering what is known about them. This will require access to a database containing diverse patient information; for example employment status and living related information alongside of medically relevant information. Again, the bi-directional arrow admits the updating of new information into the database. Clearly, this raises privacy and sensitivity issues that must be considered by stakeholders of particular applications. 

    Current research in cognitive science leads to a better understanding of how  people think, solve  problems, and learn. 
It exploits what we know about how human beings perceive the world, how they reason and make judgments and how they communicate with each other \shortcite{johnson1988computer}. This work is closely linked to work concerning \emph{Theory of Mind} (ToM).  Having a ToM refers to the human ability of explaining people's behavior based on their knowledge, their beliefs and their desires as apposed to our own. The innate human ability to put oneself in someone else's place and recognize other points of view \shortcite{
frith2005theory}. This ability is considered a key aspect of human social interactions and has already been widely explored, recognizing that to build agents that interact naturally with people, they must possess some level of ToM \shortcite{scassellati2002theory}. 
An agent that possesses a sufficient level of  ToM can recognize the goals and desires of others, and more accurately react to their emotional and cognitive states, while modifying its own behavior and communication accordingly.

There are several research branches that already benefit greatly from considering the mental model within a given interaction such as Intelligent Tutoring Systems \shortcite{woolf2010building,david2016sequencing,gal2015making}, medical health rehabilitation \shortcite{nooitgedagt2017coaching}, negotiation and argumentation \shortcite{bench2007argumentation} and  intention recognition \cite{vered2016online}.
These works demonstrate the importance and applicability of user mental models. However, if we look closer into how each work generates mental models, we notice that they all consider different aspects of the mind. This raises questions regarding which aspect of the mind should we consider? What kind of information would be required to build it and, more importantly, to maintain it? Finally, considerations should also be made such that the mental model used should also be computationally efficient, perhaps sometimes to be used in real-time interactions. Little about each of these aspects is yet known.

\vspace{-2pt}

\paragraph{Personalized Explanation Generator}
The aim of this component is to combine the knowledge from both previous components and generate a personalized explanation for the individual patient. Using information obtained from the patients' mental model, the component sifts through the general, domain-knowledge explanation and evaluates which information should be further emphasized to the patient and which is less relevant. At the end of this process, a tailored explanation of the MSP's expert recommendation is constructed and can now be passed on to the Explanation User Interface.

Recent work on XAI has stressed that explanations are not statements, but social interactions that involve people \shortcite{madumal2018towards}.  Therefore, when forming an explanation one must consider not only AI, but also research in philosophy, psychology and cognitive science. \citeauthor{miller2018insights} \citeyear{miller2018insights} argues that XAI must be \emph{context specific}. Since \emph{individual} people form part of that context XAI needs to be personalized with regards to each recipient. Research has further shown that people provide and consume explanations best when those explanations \emph{contrast} the event in question to other possible events that did not occur \shortcite{miller2018insights}. By establishing a mental model of the individual, we could narrow down the infinite contrastive possibilities to ones that are important and relevant to the individual. For example,  explaining the prescription of one course of treatment while contrasting the different advantages and disadvantages to another course of treatment in which the patient has had previous personal experience.

\paragraph{Explanation User Interface}
This component addresses the communication of the personalized explanation to either MSP, patient or both. Due to the role of the IA as a facilitator in establishing a \emph{personalized} dialogue between providers and consumers the way either consumer or provider interact with the IA is of vital importance. 
The information needs to be conveyed in a manner such that the user is not overwhelmed on one hand or left with too many questions on the other so as to  encourage trust and increase usability.
This requires considering several aspects of user interaction design such as visual vs.\ text explanations, aggregating information, whether a dialogue should be used, etc.
The MSP may also choose to use the IA as am explanatory aid. The ability for a patient to see an output introduces additional design challenges. An interface for a medical specialist may have vastly different requirements to an interface that the specialist shares with a patient and their family.

Apart from the challenge of 
making the user interaction as simple and efficient as possible
, additional challenges should be considered such as  designing an interaction system to enhance trust and acceptance by altering graphics design, content design, structure design or social cue design \shortcite{wang2005overview}. Adaptive user interfaces have also been commonly used to facilitate smoother human-computer interaction
. An adaptive or user-driven system is one that adapts to the user's specific needs and context. The adaptation may involve a different design of interface, opening a dialogue between the operator and IA, or a different representation of the systems' knowledge \shortcite{riascos2017adaptive}.

Attention should also be given to 
 recent work concerning developing and evaluating 
explanation systems \shortcite{dodge2018experts}. An analysis carried out in \citeyear{abdul2018trends} demonstrates that the streams of research in explainable systems in the AI and ML communities and in the HCI community tend to be relatively isolated \shortcite{abdul2018trends}. They have begun to address the problem by setting out a separate HCI research agenda for explainable systems that analyzes the central research clusters and how they each relate to each other. 

\vspace{-8pt}

\section{Discussion}

 We introduced the problem of personalizing inter-human communication for building trust between MSPs and patients. While we targeted the medical domain, 
 the ideas apply to any domain in which a service provider communicates to a group of highly-diverse individuals and the manner in which information is conveyed is of great importance. 
We believe that part of the solution comes from AI, and the rest requires collaboration at the intersection of several complementary research domains. We presented  a four-part conceptual framework design for a Personalized Social Interaction IA and related the advances made towards each of the components and what has yet to be done. We challenge the artificial intelligence community to address the personalization of  inter-human social communication as a means of cultivating trust and reliance between service providers and consumers.

\bibliographystyle{ACM-Reference-Format}
\bibliography{sample.bib}

\end{document}